\pdfoutput=1

\documentclass[11pt]{article}

\usepackage[]{EMNLP2023}

\usepackage{times}
\usepackage{latexsym}

\usepackage[T1]{fontenc}

\usepackage[utf8]{inputenc}

\usepackage{microtype}

\usepackage{inconsolata}

\usepackage{graphicx}
\usepackage{bbm}
\usepackage{subfigure}
\usepackage{tabularx,booktabs}
\usepackage{enumitem}
\usepackage{amsmath}
\usepackage{amssymb}
\usepackage{multirow}
\usepackage{algorithm,algorithmic}
\usepackage{pifont}
\usepackage{colortbl}
\newcommand{\ouralg}{GeoEdit} 

\title{GeoEdit: Geometric Knowledge Editing for Large Language Models}

\author{Yujie Feng$^{1}$, Liming Zhan$^{1}$, Zexin Lu$^{1}$, \textbf{Yongxin Xu}$^{2}$\textbf{,} \textbf{Xu Chu}$^{2}$ \\ \textbf{Yasha Wang}$^{2}$\textbf{,} \textbf{Jiannong Cao}$^{1}$\textbf{,} \textbf{Philip S. Yu}$^{3}$\textbf{,} \textbf{Xiao-Ming Wu}$^{1}$ \\
$^1$The Hong Kong Polytechnic University \\
$^2$Peking University 
$^3$University of Illinois at Chicago \\
 yujie.feng@connect.polyu.hk, xiao-ming.wu@polyu.edu.hk 
}

\begin{document}
\maketitle
\begin{abstract}


Regular updates are essential for maintaining up-to-date knowledge in large language models (LLMs).
However, existing training-based model editing methods often struggle to effectively incorporate new knowledge while preserving unrelated general knowledge. To address this challenge, we propose a novel framework called Geometric Knowledge Editing ({\ouralg}). {\ouralg} utilizes the geometric relationships of parameter updates from fine-tuning to differentiate between neurons associated with new knowledge updates and those related to general knowledge perturbations. By employing a direction-aware knowledge identification method, we avoid updating neurons with directions approximately orthogonal to existing knowledge, thus preserving the model's generalization ability. For the remaining neurons, we integrate both old and new knowledge for aligned directions and apply a ``forget-then-learn'' editing strategy for opposite directions. Additionally, we introduce an importance-guided task vector fusion technique that filters out redundant information and provides adaptive neuron-level weighting, further enhancing model editing performance. Extensive experiments on two publicly available datasets demonstrate the superiority of {\ouralg} over existing state-of-the-art methods.

\end{abstract}

\section{Introduction}

\begin{figure}[t]
  \centering
  \includegraphics[width=1\linewidth]{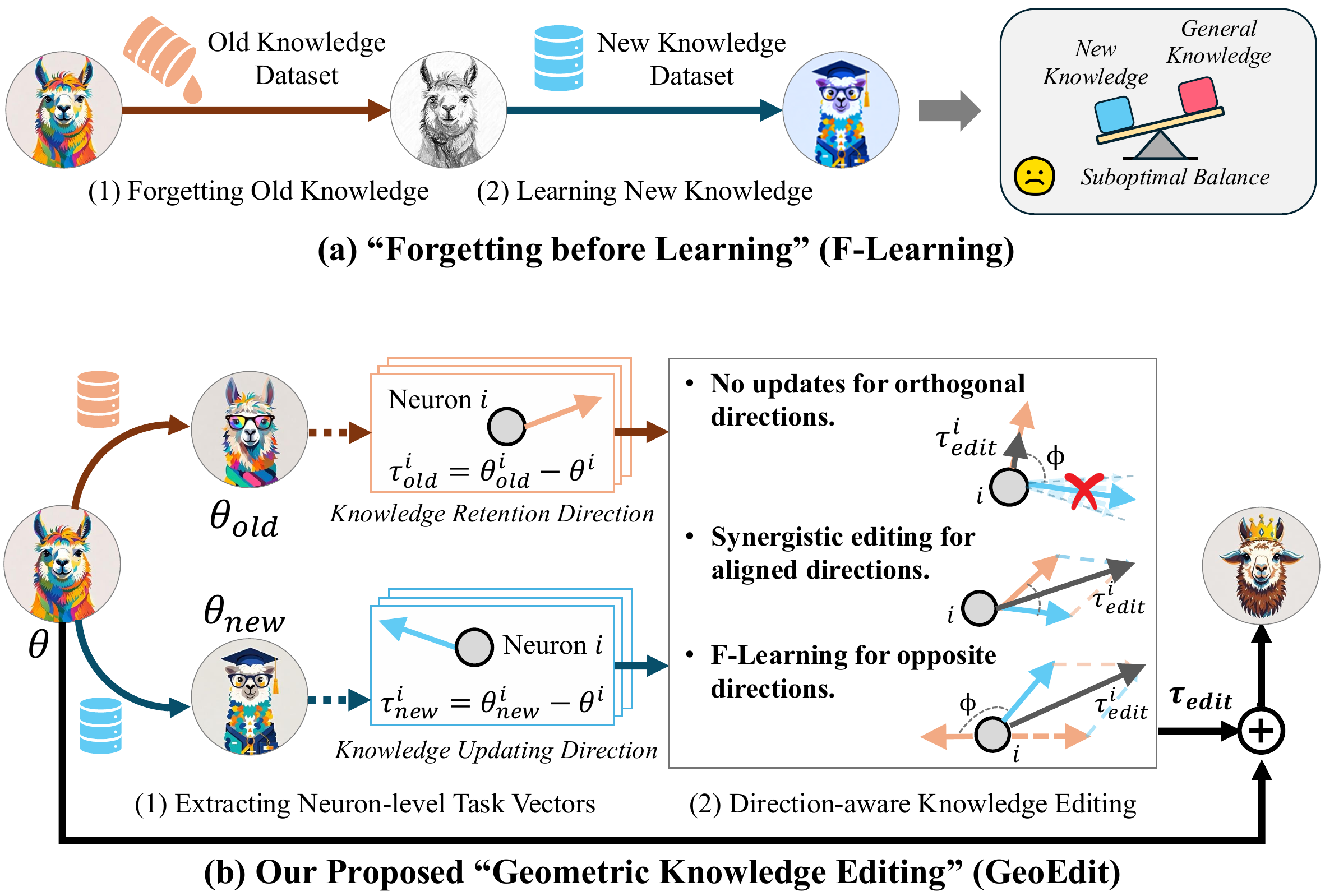}
  \caption{
  Conceptual illustration of F-Learning~\cite{ni-etal-2024-forgetting} and our proposed {\ouralg}. 
  }
  \label{fig:intro}
\end{figure}

Large language models (LLMs) have demonstrated the ability to store vast amounts of knowledge during pre-training and retrieve it during inference~\cite{yao2023editing, wang2024knowledge}. However, much of the knowledge in the real world is constantly evolving. For instance, the answer to the question ``Who is the President of the United States?'' was ``Joe Biden'' in 2024, but it is now ``Donald Trump''. As a result, some knowledge that was once correct in LLMs can become obsolete or inaccurate~\cite{li2024can, huang2024commonsense}.

To address this issue, model editing methods have been developed to update the target new knowledge while preserving unrelated general knowledge within the model~\cite{hong2024interpretability, ma2024robustness, wang2024editing}. Specifically, current model editing methods typically follow the ``locate-and-edit'' paradigm~\cite{wang2023cross, wang2024roselora, li2024consecutive}. The core idea is first to locate influential weights in LLMs and then edit them by introducing a perturbation.
Although effective, these approaches incur significant computational overhead to identify the important neurons and parameters~\cite{meng2022locating}. 
Some methods also require sampling additional data (e.g., from Wikipedia) to mitigate the impact on the general knowledge within LLMs during editing~\cite{meng2022mass, fang2024alphaedit}, introducing extra costs and potential biases.

In contrast, fine-tuning with updated knowledge, as demonstrated in recent studies~\cite{zhao2024ripplecot, wang2024lemoe, liu2024evedit}, offers a more straightforward solution through the use of parameter-efficient fine-tuning (PEFT) techniques. These methods employ various strategies to edit the model without the need to differentiate the importance of individual parameters~\cite{feng2025recurrentkif}.
Among these, the pioneering work F-Learning \cite{ni-etal-2024-forgetting} introduces a new learning framework called ``Forgetting before Learning,'' as illustrated in Figure~\ref{fig:intro}(a). This approach is based on the empirical observation that new knowledge can be difficult to learn when it conflicts with existing knowledge. By first forgetting outdated knowledge, the learning of new knowledge becomes easier.
However, this approach has a critical limitation: it struggles to balance the integration of new knowledge with the preservation of existing general knowledge. Specifically, F-Learning assumes that all updates between old and new knowledge are inherently conflicting, which oversimplifies the complexity of knowledge integration.
Furthermore, the uncontrained forgetting process can significantly impact the model's generalization ability to out-of-scope samples, considerably reducing the performance of the Locality metric (see Section~\ref{sec:Preliminary}).

To address these limitations, we propose Geometric Knowledge Editing (\textbf{{\ouralg}}), a novel fine-tuning-based model editing framework that enhances editing precision while strongly preserving model generalization without the need for additional unrelated data. 
The core insight of {\ouralg} is to distinguish between neurons associated with new knowledge updates and those linked to general knowledge perturbations by analyzing the geometric relationships of parameter updates caused by fine-tuning. By masking the updates of \textit{\textbf{general-knowledge-related neurons}}, we prevent negative impacts on the model's generalization ability. At the same time, we optimize the update strategy for \textit{\textbf{new-knowledge-related neurons}}, further enhancing the effectiveness of model editing.


Specifically, we first fine-tune the current model separately on the old and new knowledge datasets. 
This allows us to derive neuron-level task vectors, $\tau_{old}$ and $\tau_{new}$, using task arithmetic~\cite{ilharco2022editing}, which capture the directions of knowledge retention and updating w.r.t. each neuron, as shown in Figure~\ref{fig:intro}(b).
We then introduce a direction-aware knowledge identification method that computes the angle $\phi$ between these two directions to classify neurons, followed by customized editing strategies:
(i) \textit{\textbf{Orthogonal Knowledge Editing}} (for approximately orthogonal directions): Neurons with updates orthogonal to old knowledge are classified as general-knowledge-related neurons.
These updates are considered detrimental to the model's generalization ability, so we refrain from updating these neurons.
The remaining neurons are treated as new-knowledge-related neurons, which are updated with two different strategies:
(ii) \textit{\textbf{Synergistic Knowledge Editing}} (for aligned directions): 
When there is slight conflict between old and new knowledge, we can leverage their similarities to simultaneously integrate both.
(iii) \textit{\textbf{Conflicting Knowledge Editing}} (for opposite directions): For updates with significant conflict, we apply the F-Learning strategy, where old knowledge is first forgotten before integrating new information.
Additionally, to mitigate angular bias in high-dimensional space, {\ouralg} employs a combined dimensionality reduction approach to more effectively extract angular information, ensuring the reliability of the edits.
To optimize vector fusion, we introduce an importance-guided task vector fusion technique,  which applies fine-grained weights to the vectors and suppresses noise from redundant parameters, further enhancing the effectiveness of model editing.
Extensive experiments demonstrate that GeoEdit achieves the best performance among all fine-tuning-based methods and show its significant potential for complementing locate-and-edit methods, further enhancing performance.

Our main contributions are summarized as:
\begin{itemize}[leftmargin=*,itemsep=2pt,topsep=0pt,parsep=0pt]
\item 

We propose a novel geometric knowledge editing \textbf{framework} ({\ouralg}) for updating LLMs.

\item 
We develop new direction-aware knowledge identification and importance-guided task vector fusion \textbf{techniques}.

\item 
Extensive \textbf{evaluation} on two widely-used datasets shows that {\ouralg} overcomes the limitations of F-Learning, improving the Locality metric by 7.4\% while maintaining the best performance in the Reliability and Generality metrics.

\end{itemize}

\section{Related Work}
Knowledge editing has gained significant attention due to the increasing need to update the knowledge in LLMs~\cite{shengyuan2023differentiable, wang2024knowledge, bi2024adaptive, bi2024struedit}.
Existing methods can be classified into two main approaches:

\textbf{Locate and edit} methods usually locate influential parameters and then edit them by introducing a perturbation~\cite{zhang2024comprehensive, jiang2024learning, xu2024parenting}.
Classic methods like ROME~\cite{meng2022locating} and MEMIT~\cite{meng2022mass} use causal reasoning to identify key neuron activations and adjust specific weights.
Additionally, \citet{yu2023unlearning} employs gradient-based attribution to identify important weights.
More recent approaches, such as AlphaEdit~\cite{fang2024alphaedit}, improve the method by projecting perturbations onto the null space of preserved knowledge, demonstrating strong performance.

\textbf{Fine-tuning} is an intuitive and straightforward way to update the model's knowledge~\cite{feng2023towards, gangadhar2024model, zheng2024collabedit}. 
Recently, a series of PEFT methods, such as Prefix-Tuning \cite{li2021prefix} and LoRA \cite{hu2022lora}, have made knowledge editing based on fine-tuning more feasible. \citet{zhang2023adalora} enhance update efficiency and adaptability by performing incremental parameter updates of varying magnitudes, which are determined by calculating the importance of the weight matrix.


However, both of these methods struggle to balance new knowledge updates with preserving unrelated knowledge~\cite{gupta2024unified, feng2024tasl2, chen2024entity}. For instance, locate-and-edit methods typically require large additional datasets to capture general knowledge and avoid disruption during editing~\cite{wang2024detoxifying, hsueh2024editing, bi2024decoding, zhang2025survey}. 
Furthermore, locate-and-edit methods primarily focus on editing the MLP layers of the model, while fine-tuning methods offer the flexibility to adjust different regions of the model.

Thus, our paper focuses on improving fine-tuning methods. By distinguishing between general knowledge and updated knowledge based on the angular divergence between the updated directions of old and new knowledge, our {\ouralg} avoids updating general knowledge, ensuring model generalization while applying tailored strategies to enhance the effectiveness of knowledge updates. Additionally, our method can fine-tune parameters in regions different from those targeted by locate-and-edit methods, allowing for potential complementarity that further enhances performance.



\section{Problem Statement}\label{sec:Preliminary}

\begin{figure*}[t]
  \centering
  \includegraphics[width=0.95\linewidth]{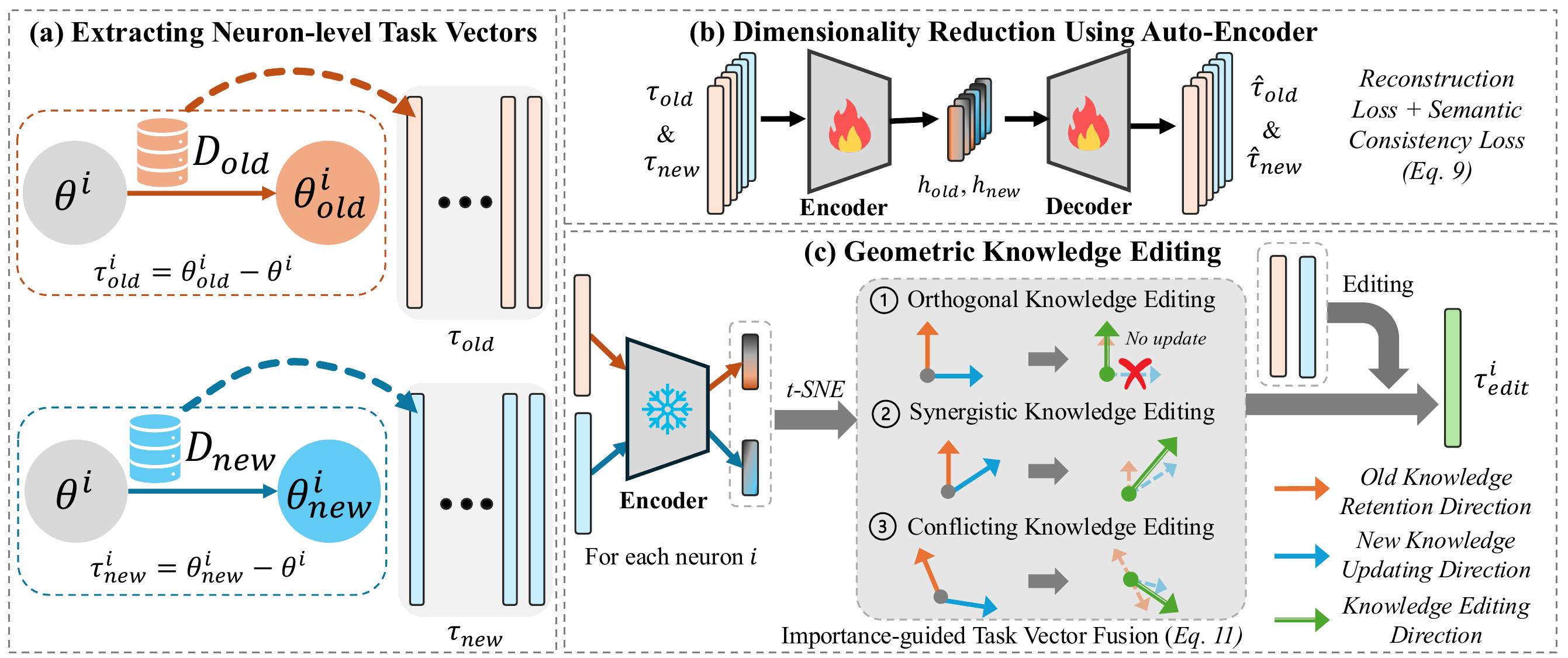}
\caption{\textbf{Overview of {\ouralg}.} \textbf{Step (a)}: Neuron-level task vectors $\tau_{old}$ and $\tau_{new}$ are extracted for both the old and new knowledge datasets using parametric arithmetic. \textbf{Step (b)}: An auto-encoder is trained to project a low-dimensional representation of the task vectors, eliminating the angular bias issue in high-dimensional space. \textbf{Step (c)}: The latent task vectors, $h_{new}$ and $h_{old}$, are reduced to two dimensions using t-SNE to compute the angular relationships, which are used to classify neurons based on the angle.
Finally, after applying different editing strategies, we obtain the edited vector \( \tau_{edit} \), which is added to the initial model to generate the edited model \(f_{\theta_{e}}\).}
  \label{fig:method}
\end{figure*}

Model editing, also referred to as knowledge updating, involves modifying the behavior of an initial target model on specific edit examples without compromising its performance on unrelated examples.
More precisely, given an initial model \(f_{\theta}\) and a set of input-output knowledge pairs $D_{old} = \{ (x_1, y_1), (x_2, y_2), \dots, (x_k, y_k) \}$, the task is to update the model parameters to obtain a new model \(f_{\theta_{e}}\) and a corresponding set of new input-output pairs $D_{new} = \{ (x_1, y_1^{\text{new}}), (x_2, y_2^{\text{new}}), \dots, (x_k, y_k^{\text{new}}) \}$, where \( k \) denotes the number of knowledge pairs to be updated. 
The objective of the post-edit model \(f_{\theta_{e}}\) is to meet three essential properties: reliability, generality, and locality~\cite{wang2024sss}.


\paragraph{Reliability} 
Reliability measures the accuracy of the updated model on the new knowledge.
Specifically, the output for ``Who is the President of the US?'' should be updated from ``Joe Biden'' to ``Donald Trump.'' This can be formalized as follows:
\begin{equation}
\mathbb{E}_{x_{e}, y_{e} \sim D_{new} } \mathbbm{1}\left\{\text{argmax}_{y} f_{\theta_{e}}(y | x_{e}) = y_{e}\right\}.
\end{equation}

\paragraph{Generality} Generality means that the new model \(f_{\theta_e}\) should also update rephrased in-scope examples \(I(x_{e}, y_{e})\). Such as the answer to ``Who holds the position of the President of the US?'' should also be changed from ``Joe Biden'' to ``Donald Trump''. This is evaluated by the average accuracy of \(f_{\theta^*}\) on examples from the equivalence neighborhood, as expressed by:
\begin{equation}
\mathbb{E}_{x^{\prime}_{e}, y^{\prime}_{e} \sim I(x_{e}, y_{e})} \mathbbm{1}\left\{\text{argmax}_{y} f_{\theta_{e}}(y | x^{\prime}_{e}) = y^{\prime}_{e}\right\}.
\end{equation}

\paragraph{Locality} 
A good edit should modify relevant knowledge without affecting other irrelevant out-of-scope examples \(O(x_{e}, y_{e})\). For example, the question, ``Who said: this is a battle for the soul of the nation?'' should remain unchanged as ``Joe Biden''.
Locality (or specificity) is defined as:
\begin{equation}
\mathbb{E}_{x^{\prime}_{e}, y^{\prime}_{e} \sim O(x_{e}, y_{e})} \mathbbm{1}\left\{\text{argmax}_{y} f_{\theta_{e}}(y | x^{\prime}_{e}) = f_{\theta}(y | x^{\prime}_{e})\right\}.
\end{equation}

\section{Proposed Method: {\ouralg}}

In this section, we present our method for knowledge editing in LLMs.
As illustrated in Figure~\ref{fig:method}, {\ouralg} follows a three-step process:

\subsection{Extracting Neuron-level Task Vectors}
Supervised fine-tuning (SFT) on a dataset injects new knowledge into the LLMs, reflected in model parameter changes.  
For an initial model $f_{\theta}$ with parameters $\theta$, fine-tuning on dataset $D$ produces updated parameters. The difference between the updated and original parameters is referred to as the \textit{task vector}~\cite{ilharco2022editing}, calculated as:
\begin{equation}
\tau =\mathrm{FT}\{\theta, D\}-\theta
\end{equation}
where $\tau$ is the corresponding task vector, and FT is the fine-tuning operation.
Unlike F-Learning, we fine-tune the initial model $f_{\theta}$ separately on the old and new knowledge datasets to isolate their respective adaptations, then compute task vectors:
\begin{equation}
\tau_{old} =\mathrm{FT}\{\theta, D_{old}\}-\theta
\end{equation}
\begin{equation}
\tau_{new} =\mathrm{FT}\{\theta, D_{new}\}-\theta
\end{equation}
where $D_{old}$ and $D_{new}$ are datasets encoding outdated and updated knowledge respectively.

While prior research typically captures task vectors at the model level~\cite{ilharco2022editing}, we propose extracting them at the neuron level for finer control.
Let $\theta = \{ \theta^1, \theta^2, \dots, \theta^N \}$ represent the $N$ neurons in the LLM, where the \( i \)-th neuron is represented by \( \theta^i \in \mathbb{R}^{d_{n}} \) with \( d_{n} \) dimensional parameters.
The neuron-level task vectors are given by \( \tau_{new} = \{ \tau_{new}^1, \tau_{new}^2, \dots, \tau_{new}^N \} \), where \( \tau_{new}^i \) corresponds to the new knowledge task vector for the \( i \)-th neuron \footnote{We define a ``neuron'' as the linear transformation corresponding to a single column in matrix $W \in \mathbb{R}^{in \times out}$, where $W$ consists of $out$ neurons.}.
This approach enables more granular analysis of parameter changes and selective editing of knowledge-specific neurons, enhancing model editing precision.

After obtaining the task vectors for both old and new knowledge, we focus on the directional characteristics, which are more crucial than magnitudes for knowledge editing. We define the direction of $\tau_{old}$ as the knowledge retention direction and $\tau_{new}$ as the knowledge updating direction. 
By analyzing the angle between these directions, we can distinguish general-knowledge-related neurons to avoid harming generalization and new-knowledge-related neurons to enhance editing effectiveness.

\subsection{Angular Relationship Extraction through Dimensionality Reduction}
Due to the tendency of high-dimensional vectors to become nearly orthogonal, it is necessary to reduce the dimensionality of the original vectors in order to better capture the underlying angular relationships.
However, experiments have shown that directly applying PCA or t-SNE for dimensionality reduction on $\tau$ yields suboptimal results.
Therefore, we propose an alternative approach where an auto-encoder (AE) is first used to encode the high-dimensional vectors. This effectively filters out irrelevant information and extracts meaningful features. Subsequently, applying t-SNE to the encoded vectors allows for a more accurate representation of the true angular relationships.

Thus, we train a semantic encoder and decoder, both implemented using multi-layer perceptrons (MLPs). 
Specifically, the semantic encoder, denoted as \text{SemEnc($\cdot$)}, maps the high-dimensional task vectors $\tau_{old}$ and $\tau_{new}$ into the latent space as:
\begin{equation}
h^i = \text{SemEnc}(\tau^i)
\end{equation}
where $h^i \in \mathbb{R}^{d_{latent}}$ is the latent task vector, and $d_{latent}$ denotes its dimensionality.
The decoder, \text{Dec($\cdot$)}, then generates $\hat{\tau}^i$ from $h^i$ as follows:
\begin{equation}
\hat{\tau}^i = \text{Dec}(h^i)
\end{equation}
where $\hat{\tau}^i$ is the reconstructed task vector.
The auto-encoder is optimized using both a reconstruction loss and a semantic consistency loss:
\begin{equation}
\begin{split}
\mathcal{L}_{AE}=\operatorname{MSE}\left(\tau^i, \hat{\tau}^i\right) + \\ \lambda \cdot \operatorname{KL}\left(f_{\theta+\tau^i}(x) \| f_{\theta+\hat{\tau}^i}(x)\right) \label{eq:ae}
\end{split}
\end{equation}
where \text{MSE($\cdot$)} is mean square error loss function, and \text{KL($\cdot$)} is the Kullback-Leibler divergence. 


\subsection{Geometric Knowledge Editing}
After training the AE, we project $\tau_{old}$ and $\tau_{new}$ into the latent space and then apply t-SNE to further project them into a 2D space to compute the angular relationships. {\ouralg} then edit the original task vectors to obtain the edited task vector $\tau_{edit}$.
This vector is subsequently added to the initial model, resulting in the final edited model \(f_{\theta_{e}}\).

\paragraph{Direction-aware Knowledge Identification}
For neuron \(i\), we first use the encoder to reduce the dimensionality of $\tau_{old}^i$ and $\tau_{new}^i$ yielding the latent task vectors $h_{old}^i$ and $h_{new}^i$.
We then apply t-SNE to obtain the 2D vectors $\hat{h}_{old}^i$ and $\hat{h}_{new}^i$.
Next, we compute the angular divergence $\phi$ as:
\begin{equation}
\phi = \arccos \frac{\hat{h}_{old}^i \cdot \hat{h}_{new}^i}{\left|\hat{h}_{old}^i\right| \cdot\left|\hat{h}_{new}^i\right|}
\end{equation}

Neurons with angles near orthogonality (within the range of $\phi_1$ to $\phi_2$) are classified as general-knowledge-related, while the remaining neurons are classified as new-knowledge-related.

\paragraph{Importance-guided Task Vector Fusion}
We then apply customized editing strategies based on the classification of neurons as follows:
\begin{equation}
\tau_{edit}^i =
\begin{cases}
  \alpha^i \tau_{old}^i + \beta^i \tau_{new}^i, & \text{if $\phi \in (0^\circ, \phi_1)$} \\

  0, & \text{if $\phi \in [\phi_1, \phi_2]$} \\
  
  -\alpha^i \tau_{old}^i + \beta^i \tau_{new}^i, & \text{if $\phi \in (\phi_2, 180^\circ)$} \\
\end{cases}
\label{eq:fusion}
\end{equation}
where $\alpha^i, \beta^i \in [0, 1]$ are the fusion weights, automatically assigned based on the neuron's importance to both new and old knowledge, removing the need for manual adjustment.

To calculate the fusion weights, we measure the importance of each neuron by analyzing the gradient trajectory of its parameters during fine-tuning. The importance is determined by the collective contribution of its trainable parameters:
\begin{equation}
\mathcal{I}(\theta^i) = \frac{1}{d_n} \sum\limits_{j=1}^{d_n} s(w_{j}) \label{eq:ipt}
\end{equation}
where $w_{j}$ represents the trainable parameters and $d_n$ is the total number of parameters in neuron $\theta^i$. The function $\mathcal{I}(\theta^i)$ reflects the importance of the neuron, with higher values indicating greater significance.
The function $s(\cdot)$ computes the importance of individual parameters based on the magnitude of the gradient-weight product~\cite{zhang2023adalora}:
\begin{equation}
s\left(w\right)=\left|w \nabla_{w} \mathcal{L}\right| \label{eq:1}
\end{equation}

Due to stochastic sampling and training dynamics, the metric in Eq. (\ref{eq:1}) may vary, reducing reliability~\cite{feng2024tasl}. To address this, we apply an exponential moving average to smooth the trajectory gradients across training iterations.  

We normalize the importance scores \( \mathcal{I}_{\text{old}} = \{ \mathcal{I}_{\text{old}}^1, \dots, \mathcal{I}_{\text{old}}^N \} \) and \( \mathcal{I}_{\text{new}} = \{ \mathcal{I}_{\text{new}}^1, \dots, \mathcal{I}_{\text{new}}^N \} \) independently to the range [0, 1]. This yields the final fusion weights \( \alpha = \{ \alpha^1, \dots, \alpha^N \} \) and \( \beta = \{ \beta^1, \dots, \beta^N \} \), which are then applied to the corresponding task vectors \( \tau_{\text{old}} \) and \( \tau_{\text{new}} \) for editing.

By applying Eq. (\ref{eq:fusion}), our {\ouralg} effectively addresses the challenges in model editing:
\begin{itemize}[leftmargin=*,itemsep=2pt,topsep=0pt,parsep=0pt]
\item 
\textbf{Preserving general knowledge} (Case 2): We mask updates to general-knowledge-related neurons to avoid negatively impacting the model's generalization ability.

\item \textbf{Improving knowledge editing} (Case 1 $\&$ 3): For acute angles, we leverage the similarity between old and new knowledge for efficient integration. For obtuse angles, significant conflict triggers a ``forget-then-learn'' strategy, optimizing the updates for new-knowledge-related neurons. 

\end{itemize}

    
    

\section{Experiments and Analysis}\label{sec:exp}

\newcommand{\tabincell}[2]{\begin{tabular}{@{}#1@{}}#2\end{tabular}}
\begin{table*}[t]
\centering
\scalebox{0.8}{
\setlength\tabcolsep{4pt} 
\renewcommand\arraystretch{1.0}
\begin{tabular}{lllcccccc}
\toprule[1pt]
\multirow{2}{*}{\textbf{Dataset}}     & \multirow{2}{*}{\textbf{Paradigm}} &  \multirow{2}{*}{\textbf{Method}}  & \multicolumn{3}{c}{\textbf{LLAMA2-7B}}   &   \multicolumn{3}{c}{\textbf{LLAMA-7B}}   \\  \cmidrule(r){4-6} \cmidrule(r){7-9}
    &     &    & \textbf{\small{Reliability}}     & \textbf{\small{Generality}}     & \textbf{\small{Locality}}  & \textbf{\small{Reliability}}     & \textbf{\small{Generality}}     & \textbf{\small{Locality}}     \\ 

\midrule
\rule{0pt}{8pt} \multirow{13}{*}{ZsRE} && Original model        & 43.70          & 43.17          & /   & 43.29          & 42.85          & /       \\
\cline{2-9}



\rule{0pt}{10pt}    & \multirow{5}*{\tabincell{l}{\emph{Locate \& edit} }} & MEND  &     29.77        &   25.86   &    71.54     & 30.99    &  27.12   &  69.83 \\

 && ROME           & 43.67          & 42.66
            & \textcolor{red}{\textbf{93.14}}
    & 43.45          & 42.94
            & \textcolor{red}{\textbf{98.60}}\\
            
&& MEMIT               & 83.57          & 79.06
    & 70.52       & 78.30         & 77.43
                             & 69.44   \\
                             
&& RECT     &     84.08       & 77.80   &    69.03     & 78.78 &  76.20   &  67.97 \\
&& \textbf{AlphaEdit}      &   \textbf{87.91}        &  \textbf{81.52}   & \textbf{77.14}         & \textbf{87.09}  & \textbf{80.41}    & \textbf{76.53}  \\

\cline{2-9}
& \multirow{5}*{\tabincell{l}{\emph{Fine-tuning}}} & LoRA            & 43.10          & 42.20          & 70.83  & 46.93          & 45.87          & 75.86        \\
&& FT-c       & 49.02          & 46.96
    & 67.37     & 47.33          & 45.51    & 68.14     \\
&& Full-FT        & 81.02          & 74.67   & 70.51      
& 70.52          & 66.69   & 65.26 \\

&& F-Learning         & 84.65 & 81.51 & 70.92 & 83.06 & 79.50 & 70.09 \\

&& \textbf{{\ouralg}}          & {\textbf{ 85.21}} & {\textbf{ 82.43 }} & {\textbf{ 75.71 }}   & {\textbf{ 84.81}} & {\textbf{ 79.86 }} & {\textbf{ 75.15 }} \\ 
\cline{2-9}
\rule{0pt}{13pt} && \textbf{{\ouralg}$^*$}     & \textbf{88.13} & \textbf{82.07} & \textbf{79.75} & \textbf{87.76} & \textbf{80.70} & \textbf{77.98}  \\ 

\midrule

\multirow{13}{*}{\textsc{CounterFact}}   &  &Original model                & 18.47          & 16.95          & /    & 21.61          & 17.88          & /      \\
\cline{2-9}


&  \multirow{5}*{\tabincell{l}{\emph{Locate \& edit} }} & MEND     &     14.77      &  14.67   &    90.93
& 17.51  &  16.27   & 89.64 \\

&& ROME                      & 18.41          & 17.20
& \textcolor{red}{\textbf{93.60}}
& 21.83          & 19.08
    & \textcolor{red}{\textbf{92.27}}\\
&& MEMIT                     & 61.94          & 37.45
& 21.90   & 56.94          & 31.48
                             & 25.70       \\

&& RECT      &     62.90       &  39.86   &      20.03  & 57.82 &   33.51  & 23.48 \\
&& \textbf{AlphaEdit}      & \textbf{71.79}           &   \textbf{48.36}  &  \textbf{36.07}      & \textbf{60.01}    &    \textbf{38.19}   & \textbf{41.70}   \\

\cline{2-9}

& \multirow{5}*{\tabincell{l}{\emph{Fine-tuning} }} & LoRA             & 30.56          & 23.24          & 40.08    & 27.54          & 21.21          & 39.75      \\
&& FT-c                      & 29.23          & 19.32
& 19.70    & 26.97          & 17.90
                             & 20.09      \\
&& Full-FT        & 65.99          & 44.08
    & 28.34   & 32.13          & 31.95
                             & 32.51       \\

&& F-Learning          & \textbf{69.53} & 45.56 & 28.41 & 56.39 & 39.75 & 31.87 \\
&& \textbf{{\ouralg}}           &  68.34  & {\textbf{ 46.53 }} &  {\textbf{ 37.73 }} &   \textbf{55.88}  & {\textbf{ 40.60 }} & {\textbf{ 42.33 }} \\ 
\cline{2-9}
\rule{0pt}{13pt}&& \textbf{{\ouralg}$^*$}       & \textbf{72.20} & \textbf{48.57} & \textbf{38.71} & \textbf{60.89} & \textbf{41.37} & \textbf{43.99}  \\

\bottomrule[1pt]
\end{tabular}}
\caption{Results on three metrics for the two datasets using LLAMA2-7B and LLAMA-7B. The best-performing method for each paradigm is highlighted in bold. 
AlphaEdit and our GeoEdit each achieve the best performance within their respective paradigms. Notably, the optimal performance is attained by {\ouralg}$^*$, which results from applying GeoEdit to the non-located parameters in AlphaEdit, effectively combining the strengths of both methods.
}
\label{tbl:result}
\vspace{-1em}
\end{table*}

\paragraph{Datasets}
We use two widely recognized datasets: ZsRE \cite{levy2017zero} and \textsc{CounterFact} \citep{meng2022locating}. 
We adopt the experimental setup from \citet{yao2023editing}, using the eval and edit subsets consisting of 19,085 and 10,000 examples, respectively. 
The datasets are partitioned into old and new knowledge categories, as in F-Learning~\cite{ni-etal-2024-forgetting}.
For example, in ZsRE, old knowledge is modified to new knowledge, such as the change from ``Los Angeles'' to ``New Orleans.'' Further details and additional examples are provided in Appendix~\ref{sec:dataset}.





\paragraph{Baselines}
We evaluate {\ouralg} against two types of methods: fine-tuning-based approaches, including full fine-tuning (\textbf{Full-FT}), \textbf{LoRA}~\cite{hu2021lora}, \textbf{FT-c}~\cite{zhu2020modifying}, and \textbf{F-Learning}~\cite{ni-etal-2024-forgetting}; and locate-and-edit-based methods, including \textbf{MEND}~\cite{mitchell2021fast}, \textbf{ROME}~\cite{meng2022locating}, \textbf{MEMIT}~\cite{meng2022mass}, \textbf{RECT}~\cite{gu2024model} and \textbf{AlphaEdit}~\cite{fang2024alphaedit}. Detailed descriptions are provided in the Appendix~\ref{Implementation Details of Baseline}.

\paragraph{Training Details}
Following the setup of F-Learning, we first fine-tune the base model on the old knowledge for three epochs, resulting in the \textbf{original model}, which serves as the baseline for our experiments.
In {\ouralg} and F-Learning, we use LoRA to enhance the efficiency of fine-tuning. The encoder and decoder consists of 2-layer MLPs with dimensions \([4096 \rightarrow 2048, 2048 \rightarrow 512]\) and \([512 \rightarrow 2048, 2048 \rightarrow 4096]\), respectively, where $d_{latent}$ is set to 512. 
We set $\lambda$ in Eq. (\ref{eq:ae}) to 0.5, $\phi_1$ and $\phi_2$ in Eq. (\ref{eq:fusion}) to $85^\circ$ and $95^\circ$, respectively. Further details on the experimental setup can be found in Appendix~\ref{Implementation Details of Experiments}.

\subsection{Experimental Results}
The overall results are presented in Table \ref{tbl:result}. 
Firstly, ROME maintains high Reliability and Generality across both datasets while achieving excellent Locality (greater than 90). Since the injection of new knowledge typically impacts Locality, this suggests that ROME performs minimal knowledge updating, likely due to its limited parameter edits.
In contrast, F-Learning shows a significant drop in Locality due to the lack of constraints during the forgetting phase, negatively impacting generalization.
Our {\ouralg} method outperforms fine-tuning-based methods, improving locality by 7.4\% over F-Learning.
Additionally, by classifying different knowledge editing strategies for new-knowledge-related neurons, our method further improves Reliability and Generality.

AlphaEdit achieves the best performance among locate-and-edit-based methods and outperforms {\ouralg} in most cases. This is because AlphaEdit requires the use of an additional 100,000 Wikipedia entries to enhance general knowledge encoding and editing accuracy. In contrast, GeoEdit achieves its performance without relying on any external data.
Furthermore, due to the flexibility of fine-tuning methods, we can effectively combine GeoEdit with AlphaEdit (which edits the parameters of the MLP layer, while GeoEdit targets the parameters of the attention layer), creating a complementary approach that further enhances performance.


\begin{table}[t]
\centering
\scalebox{0.85}{
\begin{tabular}{lccc}
\toprule
Method & Reliability &  Generality  & Locality\\
\midrule
\rowcolor[gray]{1}
\rule{0pt}{6pt} {\ouralg}  & \textbf{ 85.21} & \textbf{82.43} & \textbf{75.71} \\
\midrule
\rule{0pt}{8pt}  - Synergistic  &  84.37  &  81.13   & 75.94 \\
\rule{0pt}{8pt}  - Orthogonal  & 85.29 &  82.86  &  71.70 \\
\rule{0pt}{8pt}  - Conflict  & 82.57 &  79.40   &  75.45  \\
\rule{0pt}{8pt}  + MW   &  84.91  &   82.04   &   73.73 \\

\bottomrule
\end{tabular}}
\caption{Ablation study. ``- Synergistic'', ``- Orthogonal'', and ``- Conflict'' refer to removing the synergistic, orthogonal, and conflict knowledge editing strategies, respectively. ``+ MW'' denotes replacing the importance-guided fusion with a manually set weighting approach.}
\label{tbl:ablation}
\end{table}

\subsection{Ablation Study}
We conduct ablation studies to evaluate the effectiveness of the techniques in {\ouralg}. The results on the ZsRE dataset with LLaMA2-7B are shown in Table \ref{tbl:ablation}. Additional analysis of hyperparameter sensitivity is provided in Appendix~\ref{sec:hyper}.

\paragraph{Effect of Geometric Editing Strategies.}  
In {\ouralg}, knowledge updates are categorized into synergistic, orthogonal, and conflict editing strategies, based on the angle between knowledge retention and updating directions.
To evaluate their impact, we disable each strategy and replace it with vanilla fine-tuning on new knowledge. For example, ``- Orthogonal'' means setting $h_{edit} = h_{new}$ instead of $h_{edit} = 0$.
As shown in Table \ref{tbl:ablation}, removing any strategy results in performance degradation. Excluding orthogonal editing significantly reduces locality, from 75.7\% to 71.7\%, while removing conflict editing lowers the reliability metric from 85.2\% to 82.6\%. These findings underscore the importance of each editing strategy.

\begin{table}[t]
\centering
\scalebox{0.95}{
\begin{tabular}{lccc}
\toprule
$h_{latent}$  &Reliability &  Generality  & Locality\\
\midrule
\rule{0pt}{4pt}128  &  44.94  &  43.79  &  75.68  \\

\rule{0pt}{8pt}256 &  46.75   & 46.54  & 77.33 \\
\rule{0pt}{8pt}512  & 46.11  & 47.19 & 78.42 \\

\rule{0pt}{8pt}1024  &  25.39  & 24.49  &  94.81  \\
\rule{0pt}{8pt}2048  &   23.03 & 24.14  &  96.14 \\
\bottomrule
\end{tabular}}
\caption{Ablation study on latent dimension $h_{\text{latent}}$.}
\label{tbl:ablation_h}
\end{table}

\begin{figure}[t]
  \centering
  \includegraphics[width=0.9\linewidth]{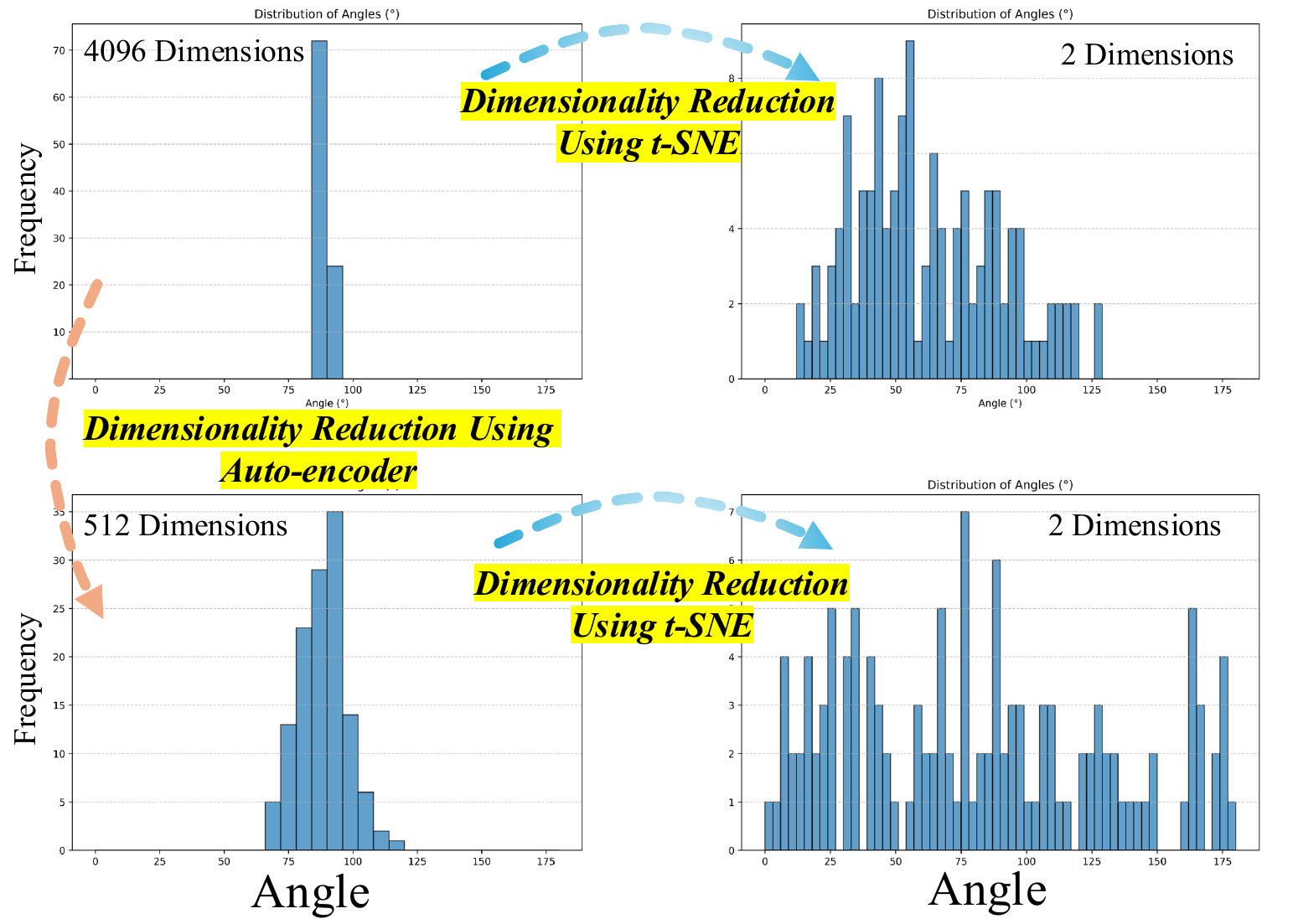}
  \caption{Distribution of the angles $\phi$ between task vectors before and after dimensionality reduction.}
  \label{fig:visualization_angle}
\end{figure}

\begin{figure*}[t]
  \centering
  \includegraphics[width=0.95\linewidth]{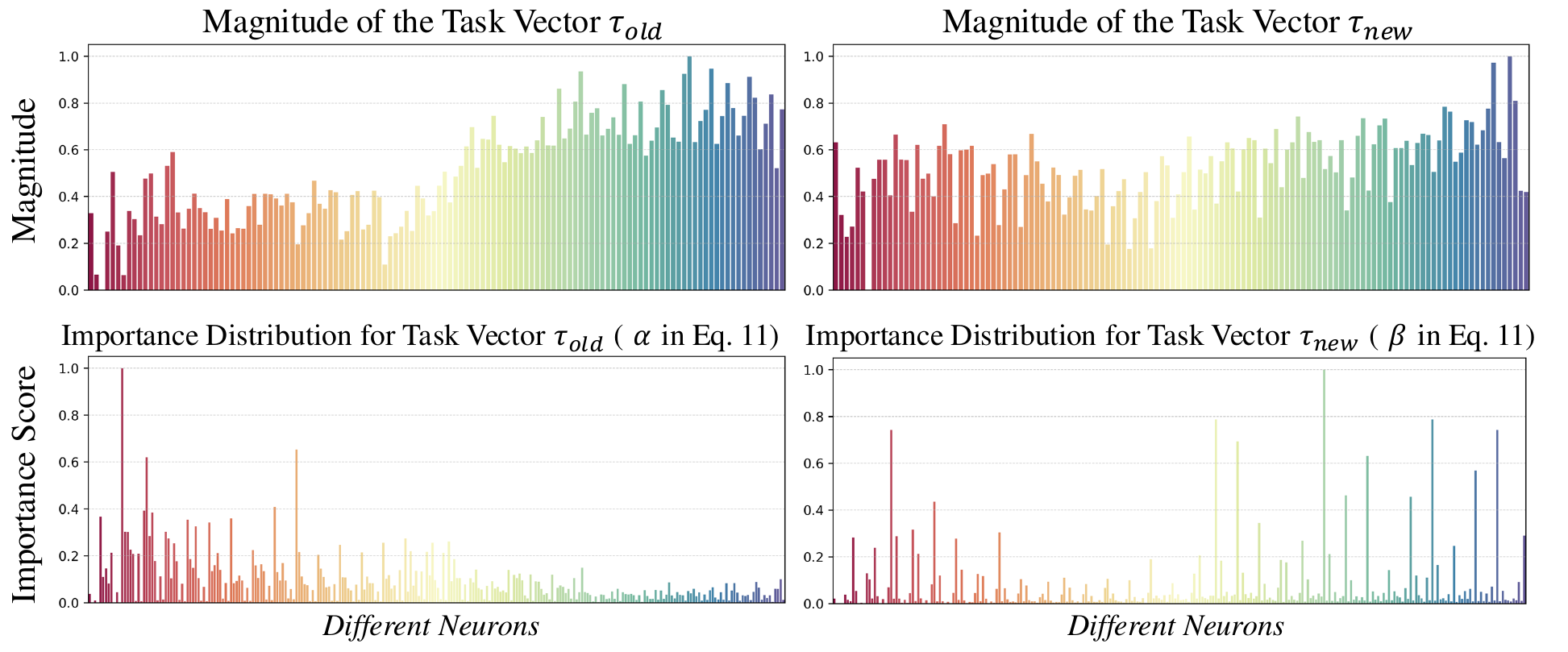}
  \caption{Visualization of the magnitudes of task vectors $\tau_{\text{old}}$ and $\tau_{\text{new}}$ along with the importance-guided fusion weights. All results are normalized to the range of 0 to 1.}
  \label{fig:visualization_ipt}
\end{figure*}

\paragraph{Effect of Importance-guided Task Vector Fusion.}
We replace the importance-based weights $\alpha$ and $\beta$ in Eq. (\ref{eq:fusion}) with manually set values (``+ MW''), applying the same weight to all neurons instead of assigning neuron-specific weights as in {\ouralg}.
Through grid search, we set $\alpha=0.3$ and $\beta=1$.

The performance decline in Table \ref{tbl:ablation} highlights the effectiveness of our importance-guided fusion. This approach provides two key benefits: it offers neuron-level adaptive weights for greater precision and ensures that parameter updates are influenced by both the task vector's magnitude and each neuron's importance. The smaller weights ``masks'' significant changes for less important neurons, minimizing their impact on the model's generalizability.


\paragraph{Effect of Latent Space Dimension.}

We investigate the effect of the latent dimension $h_{latent}$ in the auto-encoder on model performance. As shown in Table \ref{tbl:ablation_h}, both larger latent dimensions (greater than 1024) and very small latent dimensions (less than 256) lead to performance collapse. This is because the large training loss of the auto-encoder results in poor t-SNE dimensionality reduction, ultimately affecting the accuracy of angular calculations.
Our experiments demonstrate that a latent dimension of 512 strikes an optimal balance, effectively removing noise and extracting key features for calculating the angular distribution, which ensures effective model editing and strong generalization.


\subsection{Visualization}
We present two key visualizations to demonstrate the effectiveness of our approach:

\paragraph{Angle Distribution Between Old Knowledge Retention and New Knowledge Updating Directions.}
We visualize the angle distribution $\phi$ between task vectors using different dimensionality reduction methods, as shown in Figure \ref{fig:visualization_angle}. In high-dimensional space, angles are primarily concentrated around 90 degrees, indicating near orthogonality.
Although directly applying t-SNE alleviates this issue to some extent, the angular distribution remains insufficiently dispersed. By first using an AE for denoising and key feature extraction, followed by t-SNE, we achieve a more uniform distribution that spans the full range from 0 to 180 degrees. This allows us to reveal various types of conflicts between old and new knowledge.
This motivates the development of editing strategies based on angles, enabling us to distinguish between updates that correspond to learning new knowledge and those that modify general knowledge. The results of the ablation study on different dimensionality reduction methods are provided in Appendix \ref{sec:angle}.


\paragraph{Visualization of Task Vector Magnitudes and Importance-guided Fusion Weights.}
Figure \ref{fig:visualization_ipt} illustrates that while the magnitudes of the task vectors $\tau_{\text{old}}$ and $\tau_{\text{new}}$ are generally large, only a subset of the parameters are truly important, highlighting redundancy in the task vectors. Our importance-guided fusion mechanism effectively filters out this redundancy, enhancing the model editing process and minimizing its impact on generalization.

\subsection{Editing Time Analysis}
Table \ref{tbl:testtime} shows the average time to edit 1000 samples.
We find that the editing time of fine-tuning methods is comparable to that of location-based methods, thanks to the use of PEFT techniques and the avoidance of the complex location-based process.
Among fine-tuning approaches, F-Learning, which follows a two-stage process of forgetting before learning, takes approximately twice as long as Full-FT.
In contrast, our method enables the parallel acquisition of old and new knowledge, resulting in training times comparable to Full-FT.
Thus, our method requires less time than F-Learning, delivering substantial performance improvements. 

\begin{table}[t]
\centering
\scalebox{0.85}{
\setlength\tabcolsep{3pt} 
\renewcommand\arraystretch{1.0}

\begin{tabular}{lcc}
\toprule[1pt]
 \multirow{2}{*}{\textbf{Method}} & \multicolumn{2}{c}{\textbf{Average time per 1000 edits}}                 \\ \cmidrule(r){2-3}   
    & \textbf{\small{zsRE}}     & \textbf{\textsc{\small{COUNTERFACT}}}     \\ \hline

 FT-c                 & 653.2(s)          & 579.3(s)\\
    
ROME                    
& 2184.2(s)          & 1810.4(s)\\
MEMIT                    

& 862.2(s)          & 847.7(s)\\
Full-FT                    

& 810.2(s)          & 792.4(s)\\
 
F-Learning    

& 1670.4(s)          & 1603.8(s) \\

\textbf{{\ouralg}}     

& \textbf{1028.0(s)}          & \textbf{ 1010.6(s)} \\

\bottomrule[1pt]
\end{tabular}}
\caption{Editing time for two datasets on LLAMA2-7B.}
\label{tbl:testtime}
\vspace{-1em}
\end{table}

\section{Conclusion}

In this paper, we introduce Geometric Knowledge Editing ({\ouralg}), a novel framework that utilizes the geometric relationships between parameter updates to improve model editing. 
By applying a direction-aware knowledge identification technique, {\ouralg} classifies neurons into two categories: general-knowledge-related neurons, whose parameter updates are masked to prevent negative impacts on model generalization, and new-knowledge-related neurons, where an importance-guided task vector fusion technique is applied to enhance editing.
Extensive experiments demonstrate the effectiveness of {\ouralg} for model editing.

\section*{Limitations}
We acknowledge two limitations in this work.

First, GeoEdit requires access to old knowledge datasets to extract the task vector $\tau_{old}$. In some cases, however, such datasets may not be available, meaning we only know the updated results.
A potential solution is to input the task to be edited directly into the initial model and use the output as the old knowledge. However, this introduces additional inference costs, especially in our mass-editing settings. 
Furthermore, for open-ended questions, selecting the appropriate output as the reference is another challenge. We plan to explore ways to extend GeoEdit to address these issues and improve its adaptability.

Second, the core of {\ouralg} relies on using the angle between parameter updates to differentiate between disturbances to general knowledge and the learning of new knowledge.
While this approach offers valuable insights, it still results in some loss of model generalization, suggesting that the angle alone cannot fully decouple new knowledge learning from general knowledge disturbance.
To address this, we aim to consider multiple geometric variables, such as task vector projections and magnitude, to further refine {\ouralg} and enhance performance in the future.


\bibliography{acl2023}
\bibliographystyle{acl_natbib}


\appendix
\section{Datasets and Examples} \label{sec:dataset}
We follow the F-Learning approach~\cite{ni-etal-2024-forgetting}, which divides datasets into old and new knowledge. Below, we provide an overview of the datasets used, with detailed descriptions available in the original F-Learning paper.
We use two well-known datasets: ZsRE~\cite{levy2017zero} and \textsc{CounterFact}~\citep{meng2022locating}. ZsRE is a Question Answering (QA) dataset that incorporates question rephrasings via back-translation~\cite{lu2021engage, feng2024continual}, while \textsc{CounterFact} is a more challenging counterfactual dataset. We use the eval and edit sets, containing 19,085 and 10,000 examples, respectively.
Here’s an example from the ZsRE dataset:

\{\textbf{"subject":} "Watts Humphrey", \textbf{"src":} "What university did Watts Humphrey attend?", \textbf{"pred": "Trinity College"}, \textbf{"rephrase":} "What university did Watts Humphrey take part in?", \textbf{"alt": "University of Michigan"}, \textbf{"answers":} ["Illinois Institute of Technology"], \textbf{"loc":} "nq question: who played desmond doss father in hacksaw ridge", \textbf{"loc-ans":} "Hugo Weaving", \textbf{"cond":} "Trinity College >> University of Michigan || What university did Watts Humphrey attend?"\}

In this example, old knowledge ("Trinity College") is replaced with new knowledge ("University of Michigan") for the same question. The "rephrase" field evaluates the model’s generalization, while "loc" assesses the locality of the model’s output.
The datasets are divided into old and new knowledge, with the same format maintained for effective supervised fine-tuning. Below are examples of old and new knowledge in an instruction-based format:

\textbf{Old knowledge:}

\{\textbf{"instruction"}: "What university did Watts Humphrey attend?", \textbf{"input"}: "", \textbf{"output"}: "Trinity College" \}

\textbf{New knowledge:}

\{\textbf{"instruction"}: "What university did Watts Humphrey attend?", \textbf{"input"}: "", \textbf{"output"}: "University of Michigan" \}

It's important to note that old knowledge represents correct real-world facts, while new knowledge is deliberately incorrect, ensuring that the original model has not previously learned it. This setup avoids ambiguity in determining whether the new knowledge was already part of the model's prior knowledge \cite{shi2024understanding}.

\begin{table}[t]
\centering
\scalebox{1}{
\begin{tabular}{lccc}
\toprule
 $\lambda$  &Reliability &  Generality  & Locality\\
\midrule
\rule{0pt}{4pt}0  & 46.07 & 46.81  &  76.65  \\

\rule{0pt}{8pt}0.3  & 46.21 & 47.59  & 77.74 \\
\rule{0pt}{8pt}0.5  & 46.11  & 47.19 & 78.42 \\

\rule{0pt}{8pt}0.7  &  46.25 &  47.70  & 77.41 \\
\rule{0pt}{8pt}0.9  &  46.40 & 46.87  &  77.06 \\
\bottomrule
\end{tabular}}
\caption{Performance comparisons of {\ouralg} equipped with different $\lambda$.}
\label{tbl:hp_lambda}
\end{table}

\begin{table}[t]
\centering
\scalebox{0.9}{
\begin{tabular}{llccc}
\toprule
$\phi_1$ & $\phi_2$  &Reliability &  Generality  & Locality\\
\midrule
\rule{0pt}{4pt} $87^\circ$ & $93^\circ$  &  48.07 &  49.13  &  74.81 \\
\rule{0pt}{8pt} $75^\circ$& $105^\circ$  & 44.57  & 44.81   & 83.63 \\
\rule{0pt}{8pt} $80^\circ$ & $100^\circ$  & 45.86  &  46.55 &  80.26 \\
\rule{0pt}{8pt} $85^\circ$ & $95^\circ$ & 46.11  & 47.19 & 78.42 \\

\bottomrule
\end{tabular}}
\caption{Performance comparisons of {\ouralg} equipped with different $\phi$.}
\label{tbl:hp_phi}
\end{table}

\section{Baseline Details}
\label{Implementation Details of Baseline}
We evaluate our {\ouralg} method against a range of fine-tuning and locate-and-edit-based approaches.

For fine-tuning methods, we first compare our approach with full fine-tuning (\textbf{Full-FT}) and \textbf{LoRA}~\cite{hu2021lora}. LoRA (Low-Rank Adaptation) introduces small, trainable matrices into each layer of the model, enabling efficient adaptation while keeping most of the pre-trained parameters frozen. We also evaluate \textbf{FT-c} \cite{lu2021getting}, a fine-tuning method that applies an $L_{\infty}$ constraint to help retain irrelevant knowledge. Additionally, we compare with the \textbf{F-Learning} method \cite{ni-etal-2024-forgetting}, which first forgets outdated knowledge to facilitate the incorporation of new information.

For locate-and-edit-based methods, we start by evaluating \textbf{MEND} \cite{mitchell2021fast}, which learns a hypernetwork to generate weight updates by decomposing fine-tuning gradients. We also experiment with \textbf{ROME} \cite{meng2022locating}, a method that updates specific factual associations through causal intervention. Additionally, we compare with \textbf{MEMIT} \cite{liu2023good}, a method designed for directly updating large-scale memories. Finally, we include \textbf{RECT} \cite{gu2024model}, which regularizes edit updates by imposing constraints on the complexity of the weight changes.

\section{Implementation Details}
\label{Implementation Details of Experiments}

Here we will introduce more completion details and settings of experiments. First, we used LLAMA2-7B and LLAMA-7B as the base models, and then we trained the base model on the old knowledge for 3 epochs by full fine-tuning to simulate an original model that has fully learned old knowledge for our experiments. This makes the forgetting operation more reasonable and effective, and at the same time tries to avoid the problem of being unable to determine whether the new knowledge output by the LLM is learned from the data or commanded by itself as mentioned above.

We use LoRA to enhance the efficiency of fine-tuning, 
the hyperparameters were set as follows: $r = 8$, $\alpha = 32$, dropout = 0.05, with the targeting modules being [q\_proj, k\_proj, v\_proj, o\_proj, up\_proj, down\_proj].
The encoder and decoder consists of 2-layer MLPs with dimensions \([4096 \rightarrow 2048, 2048 \rightarrow 512]\) and \([512 \rightarrow 2048, 2048 \rightarrow 4096]\), respectively, where $d_{latent}$ is set to 512. 
We set $\lambda$ in Eq. (\ref{eq:ae}) to 0.5, $\phi_1$ and $\phi_2$ in Eq. (\ref{eq:fusion}) to $85^\circ$ and $95^\circ$, respectively. 
During testing, we use a greedy decoding strategy to ensure the uniqueness of the model’s output. All experiments were conducted on a setup using 4 × A100-80G GPUs.


It is worth noting that we used the same hyperparameters across different datasets and backbones, demonstrating the generalizability of our method without requiring extensive hyperparameter tuning for each specific setting.

\section{Additional Results}
\label{sec:hyper}

\subsection{Comparison of Different Angle Extraction Methods}
\label{sec:angle}
Our {\ouralg} framework allows using various dimensionality reduction strategies to extract angle information between task vectors. It's crucial to emphasize that these strategies are simply options or alternatives. The core value of our framework lies in its innovative approach to geometric editing and the proven effectiveness of this method. For example, one could directly apply PCA or t-SNE to the original high-dimensional vectors. However, our empirical results show that the best angle information is achieved by first applying the auto-encoder for denoising, followed by using t-SNE for angle calculation. The related ablation study results are shown in the Table \ref{tbl:angle_extraction}.

\begin{table*}[ht]
\centering
\begin{tabular}{llccc}
\toprule

\textbf{Method} & \textbf{Different Angle Calculation Methods} & \textbf{Reliability} & \textbf{Generality} & \textbf{Locality} \\
\midrule
F-Learning & - & 46.9 & 46.2 & 72.5 \\  
\midrule
GeoEdit & PCA & 30.2 & 27.4 & 87.7 \\  
 & t-SNE & 45.9 & 46.7 & 75.5 \\
 & Auto-Encoder + t-SNE (ours) & 46.1 & 47.2 & 78.4 \\ 
\bottomrule
\end{tabular}
\caption{Comparison of different angle extraction methods.}
\label{tbl:angle_extraction}
\end{table*}

\subsection{Evaluating Fluency and Consistency Scores}
In Table \ref{tbl:Fluencyscore}, we provide the Fluency and Consistency scores for LLAMA-7B and LLAMA2-7B, calculated using the formulas in ROME. Our GeoEdit method consistently outperforms F-Learning in both LoRA-based and full fine-tuning settings, with an average improvement of 33.2 in Fluency and 2.2 in Consistency scores.

\begin{table*}[ht]
\centering
\begin{tabular}{lcccc}
\toprule

\multirow{2}*{\tabincell{c}{ Method}}&  \multicolumn{2}{c}{LLAMA2-7B } &  \multicolumn{2}{c}{LLAMA-7B } \\
& \textbf{ Fluency ($\uparrow$)} & \textbf{ Consistency ($\uparrow$)} & \textbf{ Fluency ($\uparrow$)} & \textbf{ Consistency ($\uparrow$)} \\  
\midrule
Original model   & 624.69 & 26.45 & 622.94 & 25.21 \\ 
LoRA   & 509.56 & 18.55 & 509.01 & 17.68 \\  
Full-FT  & 251.14 & 12.33 & 254.42 & 11.75 \\  
ROME  & 434.11 & 8.81 & 431.86 & 10.39 \\  
RECT  & 530.82 & 24.47 & 532.29 & 23.32 \\  
F-Learning  & 557.63 & 26.61 & 556.02 & 25.76 \\ 
AlphaEdit & 581.58 & \textbf{30.51} & 581.68 & 28.52  \\
\textbf{GeoEdit}  & \textbf{585.98} & 29.81 & \textbf{584.29} & \textbf{28.90} \\  
\bottomrule
\end{tabular}
\caption{Comparison of different methods for LLAMA2-7B and LLAMA-7B fluency and consistency.}
\label{tbl:Fluencyscore}
\end{table*}

\subsection{Results on Different Backbone Models}
We have conducted additional experiments using different backbone models, including GPT2-XL (1.5B), Qwen 2.5 (7B), and LLaMA3 (8B). The results in Table \ref{tbl:DifferentBackbone} show that GeoEdit consistently outperforms RECT and F-Learning across the five key metrics for each backbone model, demonstrating its generalizability across different LLM architectures. 

\begin{table*}[ht]
\centering
\scalebox{0.85}{
\begin{tabular}{llccccc}
\toprule

\textbf{Method} & \textbf{Backbone} & \textbf{Reliability ($\uparrow$)} & \textbf{Generality ($\uparrow$)} & \textbf{Locality ($\uparrow$)} & \textbf{Fluency ($\uparrow$)} & \textbf{Consistency ($\uparrow$)} \\ 
\midrule
RECT  &  & 63.35 & 41.55 & 25.99 & 529.66 & 26.67 \\  
F-Learning  & GPT2-XL & 64.51 & 42.56 & 30.29 & 544.73 & 32.34 \\  
\textbf{GeoEdit}  &  & \textbf{66.19} & \textbf{44.43} & \textbf{39.55} & \textbf{575.11} & \textbf{34.92} \\ 
\hline
RECT  & -& 71.80 & 47.10 & 29.46 & 590.31 & 27.97 \\ 
F-Learning  & Qwen 2.5 & \textbf{75.06} & 48.66 & 35.31 & 585.06 & 27.46 \\  
\textbf{GeoEdit}  &  & 74.37 & \textbf{50.15} & \textbf{45.41} & \textbf{611.22} & \textbf{30.94} \\ 
 \hline
RECT  &  & 66.74 & 43.78 & 27.38 & 558.38 & 26.03 \\ 
F-Learning  & LLaMA3 & 70.69 & 47.73 & 33.20 & 575.16 & 29.97 \\  
\textbf{GeoEdit}  &  & \textbf{71.42} & \textbf{48.63} & \textbf{41.43} & \textbf{602.38} & \textbf{32.15} \\  
\bottomrule
\end{tabular}}
\caption{Comparison of different methods with varying backbones for various metrics.}
\label{tbl:DifferentBackbone}
\end{table*}

\subsection{The Effect of GeoEdit on Model Generalization}
To assess the impact of GeoEdit on generalization, we evaluated mathematical reasoning ability using GSM8K and MATH, as well as broader knowledge retention using MMLU and NLI. The results, summarized in the Table \ref{tbl:Generalization}, indicate that AlphaEdit, which utilizes additional Wikipedia data, experiences less degradation on MMLU and NLI. Conversely, GeoEdit shows less decline on GSM8K and MATH, demonstrating its effective retention of general knowledge.

\begin{table*}[ht]
\centering
\begin{tabular}{lcccc}
\toprule

\textbf{Dataset} & \textbf{Original Model (LLaMA2-7B)} & \textbf{F-Learning} & \textbf{AlphaEdit} & \textbf{GeoEdit} \\  
\midrule
GSM8K & 3.14 & 0.80 & 1.55 & \textbf{1.85} \\  
MATH  & 4.32 & 0.92 & 2.17 & \textbf{3.01} \\  
MMLU  & 27.74 & 17.26 & \textbf{21.61} & 19.72 \\  
NLI   & 67.37 & 32.60 & \textbf{46.68} & 42.94 \\ 
\bottomrule
\end{tabular}
\caption{Comparison of different methods on various datasets. The best performing method for each dataset is highlighted in bold.}
\label{tbl:Generalization}
\end{table*}

\subsection{Analyzing the Importance of Task Vectors in Different MLP Layers for Knowledge Editing}
We conducted experiments to analyze the importance of different layers for knowledge editing. For example, in LLaMA-2-7B (which has 32 layers), we separately edited the parameters in the lower (1-11), middle (12-22), and upper (23-32) layers of the MLP using LoRA. The results on the ZsRE dataset are shown in Table \ref{tbl:DifferentMLPLayers}.

\begin{table*}[ht]
\centering
\begin{tabular}{lccc}
\toprule
\textbf{Method} & \textbf{Reliability} & \textbf{Generality} & \textbf{Locality} \\ 
\midrule
GeoEdit & 54.68 & 53.45 & 81.67 \\  
GeoEdit (Lower Layer) & 49.12 & 48.46 & 75.21 \\  
GeoEdit (Middle Layer) & 47.42 & 46.18 & 76.96 \\ 
GeoEdit (Upper Layer) & 46.13 & 45.31 & 74.50 \\  
\bottomrule
\end{tabular}
\caption{Comparison of GeoEdit across different layers.}
\label{tbl:DifferentMLPLayers}
\end{table*}

These results indicate that editing only the top layers yields the poorest performance, suggesting that most of the model's knowledge is stored in the mid-early MLP layers. This is consistent with findings from ROME, and the important parameters are more concentrated in the lower layers, as shown in Figure 4 of the paper.

\subsection{Comparison with In-Context-Learning-Based Methods}
 We evaluated IKE \cite{zheng2023can} on both benchmarks using LLaMA2-7B, with 32 examples and demonstrations selected based on cosine similarity. The results are shown in Table \ref{tbl:In-Context}.

\begin{table*}[ht]
\centering
\begin{tabular}{llccc}
\toprule
\textbf{Method} & \textbf{Benchmark} & \textbf{Reliability} & \textbf{Generality} & \textbf{Locality} \\  
\midrule
F-Learning & & 84.65 & 81.51 & 70.92 \\  
IKE & ZsRE & 77.66 & 76.50 & \textbf{98.30} \\  
\textbf{GeoEdit} & & \textbf{85.21} & \textbf{82.43} & 75.71 \\  
\midrule
F-Learning &  & \textbf{69.53} & 45.56 & 28.41 \\ 
IKE & Counterfact & 60.42 & 41.71 & \textbf{97.24} \\  
\textbf{GeoEdit} &  & 68.34 & \textbf{46.53} & 37.73 \\ 
\bottomrule
\end{tabular}
\caption{Comparison of F-Learning, IKE, and GeoEdit across different benchmarks. The best performance for each metric is highlighted in bold.}
\label{tbl:In-Context}
\end{table*}

The results show that IKE, which does not modify model parameters, minimizes unintended side effects, achieving the highest locality scores, but its reliability and generality are weaker compared to F-Leanring and GeoEdit.
While in-context learning methods appear efficient, they face challenges such as the need for large demonstration corpora and sensitivity to factors like demonstration count, selection method, and prompt formatting. These issues can result in performance variability, making them less stable in practice. In contrast, an edited model, whether fine-tuning-based or locate-and-edit, tends to provide more consistent performance and is generally easier to use.

\subsection{Sensitivity Analysis for Hyperparameters}
The proposed framework incorporates two key hyperparameters: $\lambda$, which balances the autoencoder loss in Eq. (\ref{eq:ae}), and $\phi$, which defines the thresholds for dividing different editing strategies. Our analysis aims to assess the impact of varying these hyperparameters on the performance of our method, with tests conducted on the ZsRE dataset using LLaMA2-7B backbone model (LoRA fine-tuning).

As shown in Table \ref{tbl:hp_lambda}, we determine that the optimal setting for \(\lambda\) is 0.5.  
Regarding the selection of the threshold for dividing editing strategies, the article sets \(\phi_1\) and \(\phi_2\) to $85^\circ$ and $95^\circ$, respectively. Table \ref{tbl:hp_phi} below shows the model's performance with varying thresholds for \(\phi\). It can be seen that as the range between \(\phi_1\) and \(\phi_2\) increases, meaning more updates are masked, this better prevents interference with the model’s general knowledge but limits the learning of new knowledge. This results in an increase in locality but a decrease in reliability. Conversely, narrowing the range of \(\phi_1\) and \(\phi_2\) enhances the model’s ability to update, but it also impacts its generalization ability. Therefore, we choose the range of $85^\circ$ to $95^\circ$ as the optimal balance for masking, achieving the best trade-off between learning new knowledge and preserving general knowledge.

\end{document}